\documentclass[journal]{IEEEtran}
\usepackage[numbers,sort&compress]{natbib}
\usepackage{graphicx}
\usepackage{dirtytalk}
\usepackage{etoolbox}
\usepackage{booktabs}
\usepackage{multirow}

\usepackage[mathscr]{euscript}

\usepackage{caption}
\usepackage{subcaption}

\usepackage[table,xcdraw]{xcolor}
\makeatletter

\patchcmd{\@makecaption}
 {\\}
 {.\ }
 {}
 {}
 
\ifCLASSINFOpdf
\else
\fi
\usepackage{amsmath}
\usepackage{amssymb}

\usepackage{algorithm}
\usepackage{algpseudocode}
\usepackage{enumerate}
\usepackage[shortlabels]{enumitem}

\usepackage{threeparttable}
\usepackage{scalerel}
\usepackage{tikz}
\usetikzlibrary{svg.path}

\definecolor{orcidlogocol}{HTML}{A6CE39}
\tikzset{
    orcidlogo/.pic={
        \fill[orcidlogocol] svg{M256,128c0,70.7-57.3,128-128,128C57.3,256,0,198.7,0,128C0,57.3,57.3,0,128,0C198.7,0,256,57.3,256,128z};
        \fill[white] svg{M86.3,186.2H70.9V79.1h15.4v48.4V186.2z}
        svg{M108.9,79.1h41.6c39.6,0,57,28.3,57,53.6c0,27.5-21.5,53.6-56.8,53.6h-41.8V79.1z M124.3,172.4h24.5c34.9,0,42.9-26.5,42.9-39.7c0-21.5-13.7-39.7-43.7-39.7h-23.7V172.4z}
        svg{M88.7,56.8c0,5.5-4.5,10.1-10.1,10.1c-5.6,0-10.1-4.6-10.1-10.1c0-5.6,4.5-10.1,10.1-10.1C84.2,46.7,88.7,51.3,88.7,56.8z};
    }
}

\newcommand\orcidicon[1]{\href{https://orcid.org/#1}{\mbox{\scalerel*{
                \begin{tikzpicture}[yscale=-1,transform shape]
                \pic{orcidlogo};
                \end{tikzpicture}
            }{|}}}}

\usepackage{hyperref}

\begin{document}

\title{Towards Robust Plant Disease Diagnosis with Hard-sample Re-mining Strategy}

\author{\IEEEauthorblockN{Quan Huu Cap\IEEEauthorrefmark{1}\IEEEauthorrefmark{2}, Atsushi Fukuda\IEEEauthorrefmark{2}, Satoshi Kagiwada\IEEEauthorrefmark{3}, Hiroyuki Uga\IEEEauthorrefmark{4}, Nobusuke Iwasaki\IEEEauthorrefmark{5}, and Hitoshi Iyatomi\IEEEauthorrefmark{1}}

\IEEEauthorblockA{\IEEEauthorrefmark{1}Applied Informatics, Graduate School of Science and Engineering, Hosei University, Japan\\}
\IEEEauthorblockA{\IEEEauthorrefmark{2}AI Development Department, Aillis, Inc., Japan\\}

\IEEEauthorblockA{\IEEEauthorrefmark{3}Clinical Plant Science, Faculty of Bioscience and Applied Chemistry, Hosei University, Japan\\}
\IEEEauthorblockA{\IEEEauthorrefmark{4}Saitama Agricultural Technology Research Center, Japan\\}
\IEEEauthorblockA{\IEEEauthorrefmark{5}Institute for Agro-Environmental Sciences, NARO, Japan}
}

\maketitle
\begin{abstract}
    With rich annotation information, object detection-based automated plant disease diagnosis systems (e.g., YOLO-based systems) often provide advantages over classification-based systems (e.g., EfficientNet-based), such as the ability to detect disease locations and superior classification performance. 
One drawback of these detection systems is dealing with unannotated healthy data with no real symptoms present. 
In practice, healthy plant data appear to be very similar to many disease data. 
Thus, those models often produce mis-detected boxes on healthy images. 
In addition, labeling new data for detection models is typically time-consuming. 
Hard-sample mining (HSM) is a common technique for re-training a model by using the mis-detected boxes as new training samples. 
However, blindly selecting an arbitrary amount of hard-sample for re-training will result in the degradation of diagnostic performance for other diseases due to the high similarity between disease and healthy data. 
In this paper, we propose a simple but effective training strategy called hard-sample re-mining (HSReM), which is designed to enhance the diagnostic performance of healthy data and simultaneously improve the performance of disease data by strategically selecting hard-sample training images at an appropriate level. 
Experiments based on two practical in-field eight-class cucumber and ten-class tomato datasets (42.7K and 35.6K images) show that our HSReM training strategy leads to a substantial improvement in the overall diagnostic performance on large-scale unseen data. 
Specifically, the object detection model trained using the HSReM strategy not only achieved superior results as compared to the classification-based state-of-the-art EfficientNetV2-Large model and the original object detection model, but also outperformed the model using the HSM strategy in multiple evaluation metrics.  \\
\end{abstract}

\begin{IEEEkeywords}
automated plant disease diagnosis, object detection, hard-sample mining, cucumber, tomato.
\end{IEEEkeywords}

\section{Introduction}
    The early detection of plant disease is very crucial to prevent major production losses. 
However, identifying plant diseases is an expensive and laborious task because it requires experienced biology experts. 
In recent years, many deep learning-based techniques have been actively developed to automate plant disease classification and, thus, help farmers mitigate plant-productivity losses \citep{kawasaki2015, mohanty2016, ferentinos2018deep, ramcharan2019mobile, boulent2019convolutional, zhang2019cucumber, chen2021identification, bao2021lightweight, borhani2022deep, khan2022deep, wang2022practical, li2023novel, thai2023formerleaf}. 
They have achieved remarkable results in the diagnosis of plant diseases in various crops, ranging from cucumber \citep{kawasaki2015, zhang2019cucumber, wang2022practical}, cassava \citep{ramcharan2019mobile, thai2023formerleaf}, rice \citep{chen2021identification, borhani2022deep}, and wheat \citep{bao2021lightweight, borhani2022deep} to grape \citep{boulent2019convolutional} and apple \citep{khan2022deep}. 
These classification-based systems typically accept input in the form of a narrow-range image, with one or a few targets located in the center. 
Despite excellent diagnostic capabilities, these systems still have limitations in the classification of wide-range images, which can contain multiple targets or even dozens of targets \citep{cap2018}. 

To resolve the problem, disease detection methods based on general object detection models, such as the SSD \citep{liu2016ssd}, Faster R-CNN \citep{ren2016faster}, or YOLO families \citep{redmon2016you, redmon2017yolo9000, redmon2018yolov3, bochkovskiy2020yolov4, jocheryolov5} have been proposed \citep{fuentes2017robust, ozguven2019automatic, liu2020tomato, kim2021improved, qi2022improved, chen2022plant, xue2023yolo}. 
Taking practicality into consideration, detection-based disease diagnosis systems typically have more advantages than classification-based systems. 
First, detection-based systems not only can predict but also provide the location of diseases, thus enriching the diagnostic information. 
Second, these systems are capable of handling wide-range images and exhibit greater robustness to changes in distance between cameras and images as compared to classification-based systems. 
Third, detection-based systems can be treated as classification models based on the highest confidence bounding box if assuming only one disease appears in each input image (i.e., no multiple infections). 
As we experienced, their robustness against the covariate shift problem, i.e., performance is high for test data derived from the same imaging environment as used for training, but much lower for data derived from a different environment \citep{cap2018, suwa2019comparable, saikawa2019aop, shibuya2022validation, wayama2023} is normally superior to that of classification-based systems. 

Despite their advantages, an issue arises for these systems when dealing with unannotated healthy data. 
In practice, detection-based systems should have the ability to detect healthy cases. 
However, because there are no real symptoms available, the question of where to put the annotation data on healthy images is confusing. 
To our knowledge, there have been no detection-based plant disease diagnosis studies that included healthy data ever proposed. 
A possible way to resolve this issue might involve annotating bounding boxes for the healthy data. 
Nevertheless, the number of healthy leaves often vastly outnumbers the diseased leaves, thus making the process of labeling laborious and time-consuming. 

\begin{figure*}[!t]
\centering
\begin{subfigure}[b]{0.35\textwidth}
    \centering
    \includegraphics[width=\textwidth]{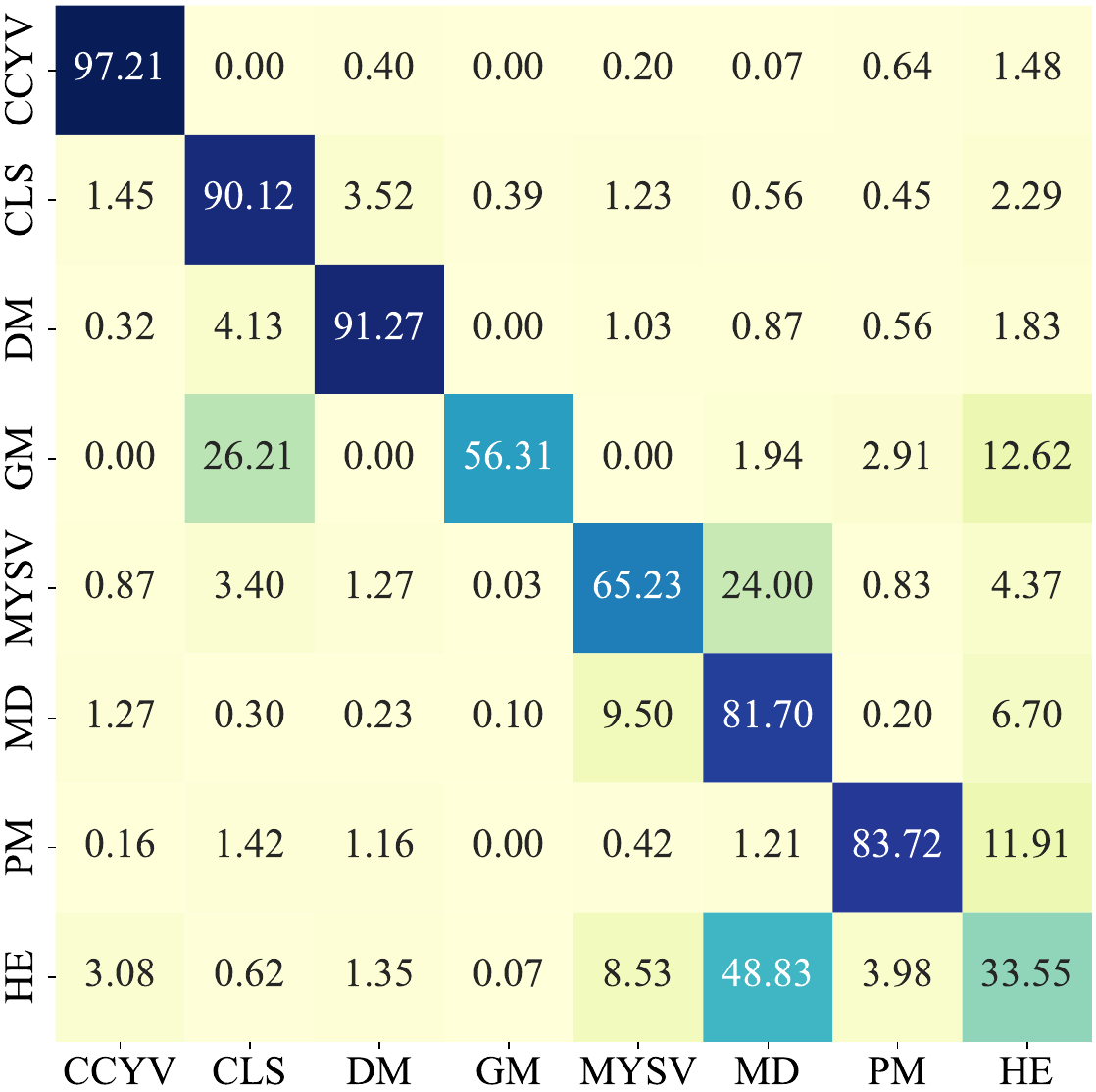}
    \caption{Naive detection model}
\end{subfigure}
\begin{subfigure}[b]{0.35\textwidth}
    \centering
    \includegraphics[width=\textwidth]{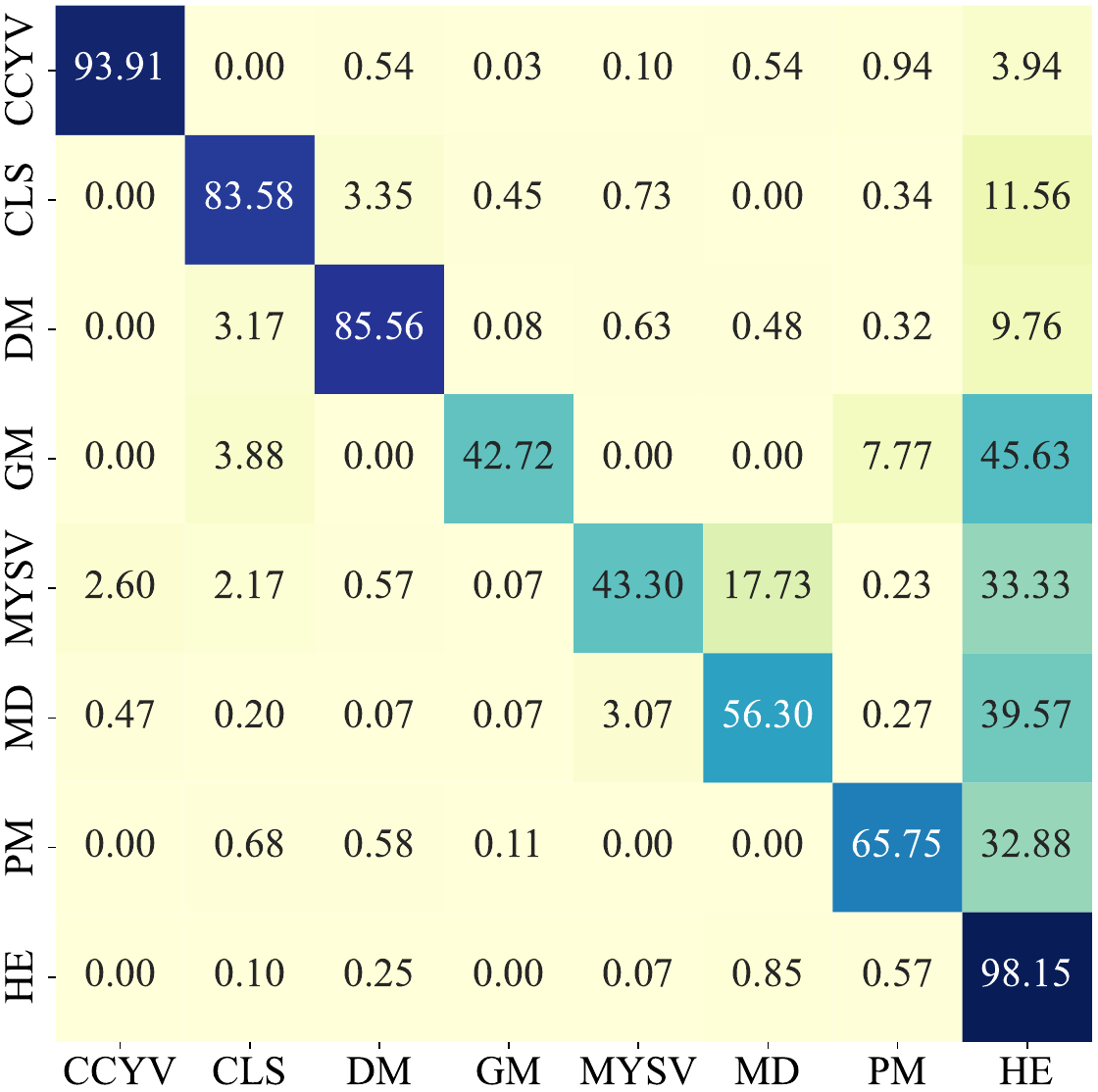}
    \caption{Detection model with HSM strategy}
\end{subfigure}
\caption{
    The comparison between the naive approach (a) and the hard-sample mining (HSM) strategy (b) on the cucumber dataset in recall (confusion matrices normalized over rows).
}
\label{fig:fig_1}
\end{figure*}
To mitigate this issue, we considered several potential approaches. The first general solution is to create detection rules. 
For example, if no bounding boxes are detected, an input image will be considered to belong to the correctly predicted healthy class. 
However, we experienced that this naive strategy does not work well, because the detection model may falsely predict healthy as diseased since some diseases have appearances very similar to healthy. 
Fig. \ref{fig:fig_1}(a) shows the confusion matrix of a YOLOv5-based cucumber disease diagnosis model, with this naive strategy being trained on the dataset shown in Table \ref{tab:table_1} (in section \ref{sec:sec_2_1}). 
Note that the training images (Dataset A) and evaluation images (Dataset B) are taken from different farm fields. 
In this case, the model mis-diagnosed nearly half of the healthy (HE) cases as mosaic viral diseases (MD). 

The second approach is applying a two-stage system that combines a disease detection system with a healthy classifier. 
Once the input is classified as unhealthy, the disease detection model will be applied. 
Although this approach seems reasonable, it will be shown later that this is not an optimal solution, because the ultimate performance heavily depends on the first stage and this approach is also quite cumbersome. 

The last and most promising approach, in our opinion, is utilizing the hard-sample mining (HSM) technique \citep{shrivastava2016training} on healthy data. 
The HSM technique is a popular training strategy aimed at boosting the detection performance for \say{hard-to-detect} bounding boxes. 
This strategy has been used in many applications, ranging from person re-identification \citep{smirnov2018hard, chen2020hard, wang2020joint, han2021hmmn}, speech recognition \citep{xue2019hard, wang2022mining, hu2022synthetic}, medical diagnosis \citep{li2019deep, tang2019uldor, wang2021retinal, okamoto2021gastric, zhu2022iternet++} to the agricultural field \citep{zhang2020recognition, you2021deep, kim2021improved, lin2022boosted, dai2022detection}. 
When using a detection model trained on annotated disease data, healthy images that were mis-predicted as diseases (i.e., false positives) will be considered healthy hard-samples. 
This HSM strategy is expected to greatly reduce the labeling process by leveraging the mis-detected bounding boxes as the annotation data for model re-training. 

\begin{figure*}[!t]
\centering
\includegraphics[width=0.95\textwidth]{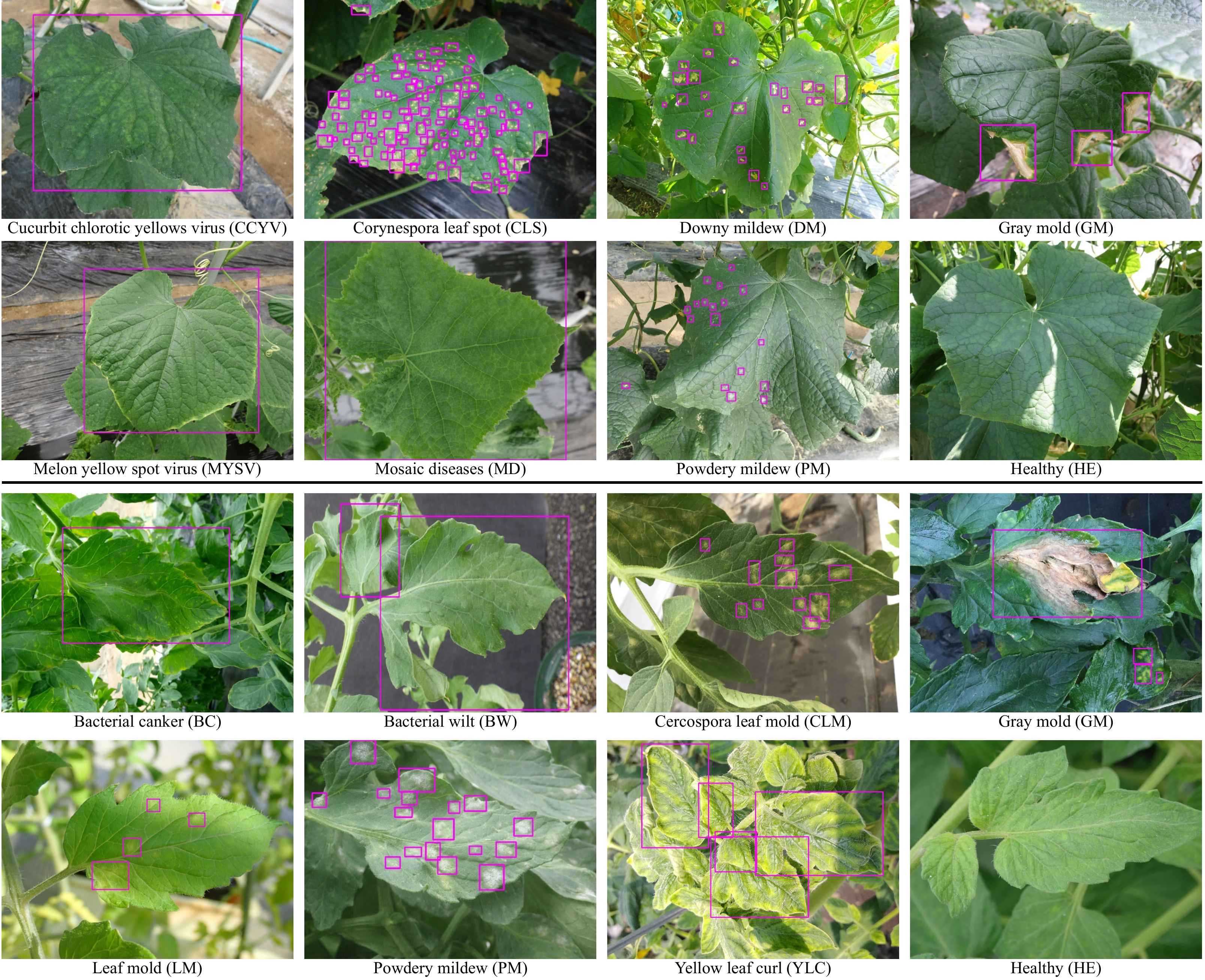}
\caption{Annotated leaf disease images, along with healthy leaves in our two datasets (first two rows: cucumber data, last two rows: tomato data). Images of some disease classes show faint disease symptoms and are difficult to distinguish from healthy classes.}
\label{fig:fig_2}
\end{figure*}

In reality, however, some diseases are very similar to healthy cases, and the two are thus difficult to distinguish, especially given early disease symptoms (see Fig. \ref{fig:fig_2}). 
We found that adding many healthy hard-samples to train the model could break the disease detection performance. 
Namely, the more hard-samples were used to train the model, the fewer false negatives (wrongly detected as healthy) that could not previously be explicitly trained there were, but the sensitivity of the disease detection itself may be reduced. 
Fig. \ref{fig:fig_1}(b) shows the confusion matrix for the cucumber detection model with the HSM strategy. 
The diagnostic performance for the healthy class (in recall) has been significantly improved, but it was severely reduced for some disease classes. 
In this case, adjusting the number of hard-samples used for training is necessary to achieve practical performance. 

In this paper, we propose a simple but effective training strategy called hard-sample re-mining (HSReM). 
Based on the results of the HSM strategy, HSReM automatically selects the hard-sample classes that help to boost the performance for both healthy and diseased while eliminating those with negative impacts on the diagnostic performance for other diseases. 
We showed that the proposed HSReM strategy is effective for practical large-scale cucumber and tomato datasets in a disease classification task with realistic and strict evaluation. 
Our contributions can be summarized as follows: 
\begin{itemize}
    \item We introduce HSReM as a novel training strategy to effectively enhance the fine-grained diagnostic performance of detection-based models when diseases and healthy data appear with high similarities. 
    \item We provide comprehensive empirical validation of the proposed HSReM strategy with prohibitively strict criteria, wherein test images were collected in distinct farm fields from the training images. 
    \item Our results demonstrate that object detection models trained using the HSReM strategy not only surpassed the classification-based state-of-the-art EfficientNetV2-Large model \citep{tan2021efficientnetv2} and the original object detection model, but also outperformed models using the HSM strategy in multiple evaluation metrics. 
\end{itemize}

\section{Materials and methods}
    \subsection{Dataset acquisition}\label{sec:sec_2_1}
In this paper, we use practical in-field cucumber and tomato datasets to train our disease detection models. 
Note here again that all data had only one disease class in each input image (i.e., no multiple infections), and we consider this a single-label classification problem. 
The cucumber and tomato images were obtained from 14 farms in Japan during different time periods. 
The sizes of the images are very diverse, with the smallest edge length ranging from 1,500 to 3,000 pixels. 
The collected plant images were divided into two datasets based on where they were taken: Dataset A and Dataset B. 
For rigorous evaluation, test images must be collected from outside the field where the training images were taken \citep{shibuya2022validation, wayama2023}. 
In this work, Dataset A and Dataset B are properly and strictly separated on a farm-by-farm basis. 
For both types of plants, the bounding boxes for disease symptoms on images from Dataset A were annotated by biology experts. 
For Dataset B, only class-level labels are available. 

\begin{figure*}[!t]
\centering
\includegraphics[width=0.64\textwidth]{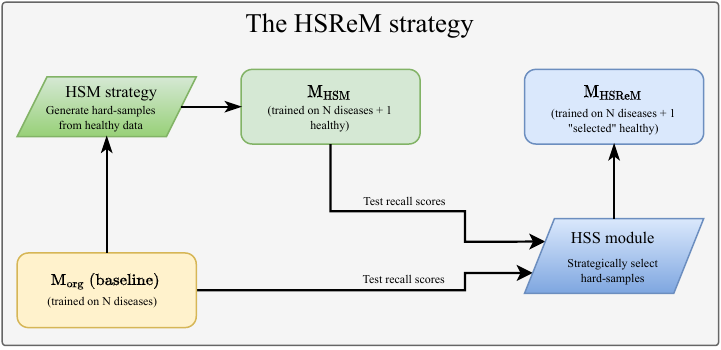}
\caption{The step-by-step visualization of the proposed HSReM strategy.}
\label{fig:fig_3}
\end{figure*}
Unlike general object detection datasets, plant disease symptoms appear in a very complicated way, with many forms and development stages. 
Therefore, it is very difficult to define the rules with which to annotate all these symptoms. 
If a leaf has many symptoms, we roughly annotate the main locations that show the strong presence of symptoms. 
Fig. \ref{fig:fig_2} illustrates some examples of annotated leaf images in our datasets. 
In total, there are 15 classes (14 disease classes and one healthy class) in both the cucumber and tomato datasets: \emph{bacterial canker} (BC), \emph{bacterial wilt} (BW), \emph{cucurbit chlorotic yellows virus} (CCYV), \emph{cercospora leaf mold} (CLM), \emph{corynespora leaf spot}  (CLS), \emph{corynespora target spot} (CTS), \emph{downy mildew} (DM), \emph{gray mold} (GM), \emph{late blight} (LB), \emph{leaf mold} (LM), \emph{melon yellow spot virus} (MYSV), \emph{mosaic diseases} (MD), \emph{powdery mildew} (PM), \emph{yellow leaf curl} (YLC), and \emph{healthy} (HE). 
Note that the annotation bounding boxes are not available for healthy data. 
The details of all classes from the two datasets are shown in Table \ref{tab:table_1}. 
\begin{table*}[!t]
\centering
\caption{Statistics of the on-site cucumber and tomato datasets}
\label{tab:table_1}
\resizebox{0.9\linewidth}{!}{
\begin{threeparttable}
\begin{tabular}{lrrrrrclrrrrr}
\cline{1-6} \cline{8-13}
\multicolumn{6}{c}{Cucumber}                                                                                                                                        &  & \multicolumn{6}{c}{Tomato}                                                                                                                    \\ \cline{1-6} \cline{8-13} 
\multirow{2}{*}{Class} & \multicolumn{4}{c}{Dataset A}                                                             & \multicolumn{1}{c}{\multirow{2}{*}{Dataset B}} &  & \multirow{2}{*}{Class} & \multicolumn{4}{c}{Dataset A}                                       & \multicolumn{1}{c}{\multirow{2}{*}{Dataset B}} \\ \cline{2-5} \cline{9-12}
                       & Train                & \# of boxes          & Test                 & \# of boxes          & \multicolumn{1}{c}{}                           &  &                        & Train           & \# of boxes     & Test           & \# of boxes    & \multicolumn{1}{c}{}                           \\ \cline{1-6} \cline{8-13} 
CCYV                   & 1,970                & 3,950                & 219                  & 465                  & 2,972                                          &  & BC                     & 1,205           & 2,418           & 133            & 270            & 257                                            \\
CLS                    & 3,231                & 99,698               & 359                  & 10,017               & 1,791                                          &  & BW                     & 1,201           & 3,043           & 133            & 362            & 425                                            \\
DM                     & 2,404                & 32,876               & 267                  & 3,465                & 1,260                                          &  & CLM                    & 2,407           & 14,914          & 267            & 1,618          & 2,602                                          \\
GM                     & 653                  & 2,490                & 72                   & 315                  & 103                                            &  & CTS                    & 1,169           & 8,387           & 129            & 746            & 1,291                                          \\
MYSV                   & 2,062                & 2,195                & 229                  & 241                  & 3,000                                          &  & GM                     & 1,257           & 1,646           & 139            & 173            & 462                                            \\
MD                     & 3,241                & 5,757                & 360                  & 638                  & 3,000                                          &  & LB                     & 1,516           & 2,468           & 168            & 287            & 549                                            \\
PM                     & 3,144                & 58,312               & 349                  & 7,362                & 1,898                                          &  & LM                     & 1,916           & 13,957          & 206            & 1,482          & 451                                            \\
HE                     & 8,000                & N/A                  & 446                  & N/A                  & 4,000                                          &  & PM                     & 1,987           & 22,909          & 220            & 2,377          & 4,247                                          \\ \cline{1-6}
\textbf{Total}         & \textbf{24,705}      & \textbf{205,278}     & \textbf{2,301}       & \textbf{22,503}      & \textbf{18,024}                                &  & YLC                    & 2,033           & 7,788           & 225            & 833            & 878                                            \\ \cline{1-6}
\multicolumn{1}{c}{}   & \multicolumn{1}{c}{} & \multicolumn{1}{c}{} & \multicolumn{1}{c}{} & \multicolumn{1}{c}{} & \multicolumn{1}{c}{}                           &  & HE                     & 7,000           & N/A             & 372            & N/A            & 2,742                                          \\ \cline{8-13} 
\multicolumn{1}{c}{}   & \multicolumn{1}{c}{} & \multicolumn{1}{c}{} & \multicolumn{1}{c}{} & \multicolumn{1}{c}{} & \multicolumn{1}{c}{}                           &  & \textbf{Total}         & \textbf{21,691} & \textbf{77,530} & \textbf{1,992} & \textbf{8,148} & \textbf{13,904}                                \\ \cline{8-13} 
\end{tabular}
\begin{tablenotes}[flushleft]
    \item[$\dag$] For rigorous evaluation, Dataset A, for training, and Dataset B, for evaluation, are strictly separated from the farms where the images were acquired.
\end{tablenotes}
\end{threeparttable}
}
\end{table*}

\subsection{The proposed hard-sample re-mining strategy}
The hard-sample re-mining (HSReM) strategy is an improved version of the HSM strategy, with the goal of sustaining or improving fine-grained diagnostic performance for not only healthy but also all diseases. 
The HSReM strategy consists of the HSM strategy, followed by a newly proposed hard-sample selection (HSS) module that strategically selects hard-sample training images at an appropriate level. 
As for the prerequisites, in this work, we define the HSM strategy as an offline re-training method. 
Given an original trained detection model $\mathrm{M_{org}}$, which is designed to detect $N$ diseases, any mis-detected samples in a novel class data are considered to be the hard-samples for that class, and they can then be used for re-training a new detection model, $\mathrm{M_{HSM}}$, on $N$ diseases + novel classes. 
\begin{figure}[!t]
\centering
\includegraphics[width=\linewidth]{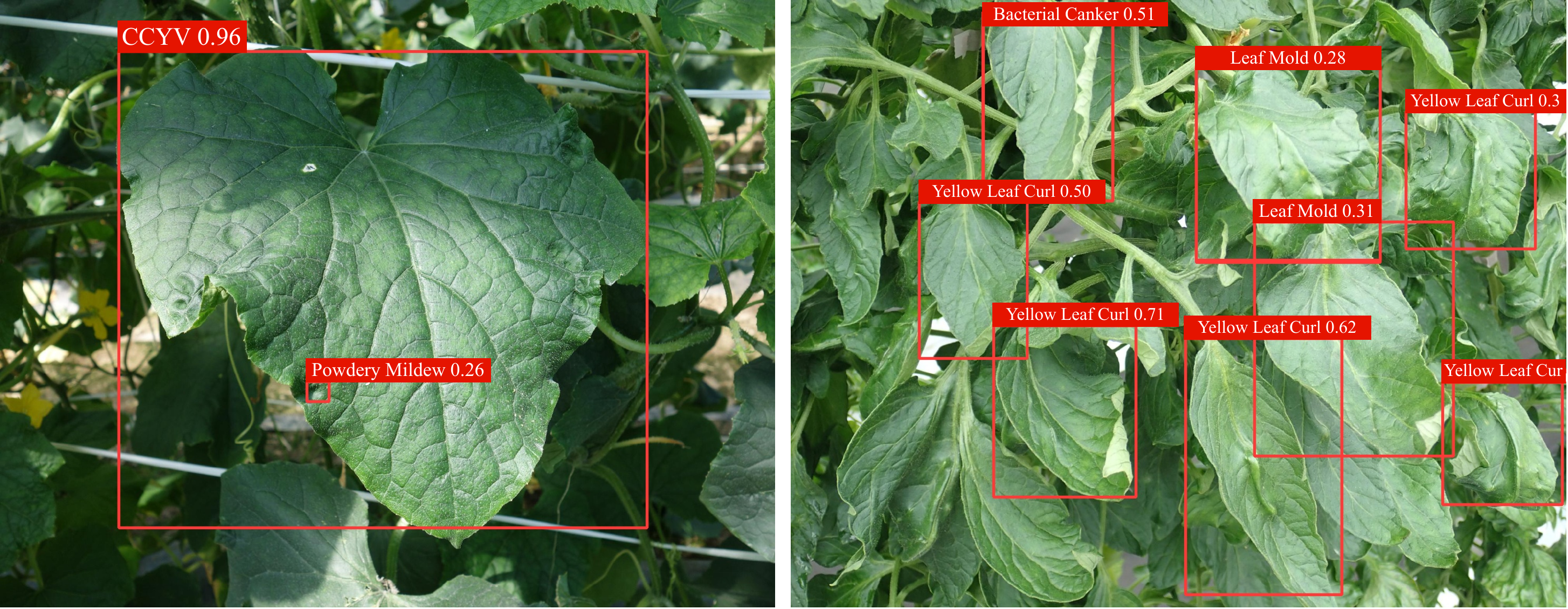}
\caption{Mis-detected bounding boxes on healthy images. These boxes are considered hard-sample bounding boxes.}
\label{fig:fig_4}
\end{figure}

Fig. \ref{fig:fig_3} illustrates our proposed HSReM strategy in a step-by-step manner. 
From the original baseline model ($\mathrm{M_{org}}$), as an intermediate step, the HSM strategy is first used to generate hard-samples (mis-detected bounding boxes) from healthy training data (Fig. \ref{fig:fig_4}) for training the detection model $\mathrm{M_{HSM}}$ on $N$ diseases + a healthy class (i.e., $N+1$ classes in total). 
Here, the HSM strategy plays an important role in probing the possibility of model performance degradation in disease classes. 
Then, the proposed HSS module strategically selects hard-samples in the appropriate amount to training the final detection model, $\mathrm{M_{HSReM}}$. 
It is worth noting that while previous studies required data to be annotated beforehand in order for them to work with HSM \citep{zhang2020recognition, you2021deep, kim2021improved, lin2022boosted, dai2022detection}, we apply HSM and HSReM differently, in an unsupervised manner, to mine the hard-samples from unannotated healthy data, which is similar to \citep{okamoto2021gastric}. 
\begin{algorithm}[!t]

\renewcommand{\algorithmicrequire}{\textbf{Inputs:}}
\renewcommand{\algorithmicensure}{\textbf{Outputs:}}

\caption{The hard-sample selection (HSS) module}\label{alg:HSS_module}

\begin{algorithmic}[1]
\Require
    \Statex - $R_\mathrm{org}[\mathrm{disease}_i]$:  recall value of $i$-th disease class of the $\mathrm{M_{org}}$ model on the test data. 
    \Statex - $R_\mathrm{HSM}[\mathrm{disease}_i]$:  recall value of $i$-th disease class of the $\mathrm{M_{HSM}}$ model on the test data. 
    \Statex - $\mathbb{D}$: a dictionary with disease classes as keys, and the values are the number of hard-sample images that mis-detected those classes. 
    \Statex - $\theta$: the pre-defined threshold.

\Ensure The updated version of $\mathbb{D}$.
    
\For {$\mathrm{disease}_i$ \textbf{in} $\mathbb{D}$}
    \If {$R_\mathrm{org}[\mathrm{disease}_i] - R_\mathrm{HSM}[\mathrm{disease}_i] > \theta$}
        \State Remove all hard-sample images in $\mathbb{D}[\mathrm{disease}_i]$
    \EndIf
\EndFor
\State \Return $\mathbb{D}$

\end{algorithmic}
\end{algorithm}

The core idea of the HSS module is that, if the diagnostic performance of a disease class has dropped by more than a pre-defined threshold $\theta$ (in percentage terms) after the HSM step, all the healthy hard-samples that correspond to that class (healthy images but mis-detected as that disease) should be removed. 
Specifically, given the two detection models $\mathrm{M_{org}}$ and $\mathrm{M_{HSM}}$, the HSS module decides which hard-samples to keep by comparing the recall scores for each class based on the test data between the two models. 
The pseudo-code for the HSS module is shown in the Algorithm \ref{alg:HSS_module}. 

In this work, the pre-defined threshold $\theta$ (in percentage terms) is determined by our preliminary experiments. 
Specifically, we applied binary search to search for the integer value $\theta$ within [lower bound, upper bound] of $[3,9]$. 
In our preliminary experiments, we found that the pre-defined threshold $\theta=6\%$ worked well for both the tomato and cucumber data. 
After applying the HSS module, the newly updated hard-sample dataset will be used to train the final detection model, $\mathrm{M_{HSReM}}$. 
The selection of an appropriate threshold $\theta$ is crucial. 
We suggest users to search for the threshold $\theta$ value in the range of $[3,\beta]$ where $\beta$ is the upper bound and $\beta=\mathrm{max}(R_\mathrm{org}[\mathrm{disease}_i] - R_\mathrm{HSM}[\mathrm{disease}_i]) \;\forall\; \mathrm{disease}_i \in \mathbb{D}$. 

\section{Experiments and results}
    We compare our proposed model with several approaches. 
Specifically, we train the following five models: 
\begin{enumerate}[label={(\arabic*)}]
    \item Baseline ($\mathrm{M_{org}}$): a baseline detection model based on the YOLOv5-L6 model \citep{jocheryolov5}. 
    \item Baseline+HSM ($\mathrm{M_{HSM}}$): a YOLOv5-L6 model trained with the HSM strategy. 
    \item Baseline+HSReM ($\mathrm{M_{HSReM}}$): a YOLOv5-L6 model trained with the HSReM strategy (\textbf{proposed model}). 
    \item CNN (EfficientNetV2): a classification model based on the EfficientNetV2-Large model \citep{tan2021efficientnetv2}. 
    \item Two-stage: a model with an EfficientNetV2-Large-based binary classifier for healthy/disease in the first stage, combined with the baseline $\mathrm{M_{org}}$ model (1) in the second stage. 
\end{enumerate}

Dataset A train is used to train all the disease diagnostic models. 
Dataset A test and Dataset B are used to evaluate the performance of the models on images from the same and different farms from which the training data was captured, respectively. 
Table \ref{tab:table_2} shows the statistics for the obtained healthy hard-samples after using the HSM and HSReM strategies for both cucumbers and tomatos (Dataset A train). 
Each row represents the number of healthy images that were mis-detected to a disease class. 
The number of hard-samples shown here were used to train the $\mathrm{M_{HSM}}$ and  $\mathrm{M_{HSReM}}$ models. 
\begin{table}[!t]
\centering
\caption{Statistics of the healthy hard-samples after HSM and HSReM strategies on cucumber and tomato datasets}
\label{tab:table_2}
\resizebox{0.9\linewidth}{!}{
\begin{threeparttable}
\begin{tabular}{lrrclrr}
\cline{1-3} \cline{5-7}
\multicolumn{3}{c}{\begin{tabular}[c]{@{}c@{}}Cucumber\\ (\#healthy=8,000)\end{tabular}} &  & \multicolumn{3}{c}{\begin{tabular}[c]{@{}c@{}}Tomato\\ (\#healthy=7,000)\end{tabular}} \\ \cline{1-3} \cline{5-7} 
\multirow{2}{*}{Class}            & \multicolumn{2}{c}{Hard-samples}                     &  & \multirow{2}{*}{Class}           & \multicolumn{2}{c}{Hard-samples}                    \\ \cline{2-3} \cline{6-7} 
                                  & HSM                       & HSReM                    &  &                                  & HSM                      & HSReM                    \\ \cline{1-3} \cline{5-7} 
CCYV                              & 1,003                     & 1,003                    &  & BC                               & 595                      & 595                      \\
CLS                               & 194                       & 0                        &  & BW                               & 931                      & 0                        \\
DM                                & 307                       & 307                      &  & CLM                              & 64                       & 64                       \\
GM                                & 24                        & 0                        &  & CTS                              & 50                       & 50                       \\
MYSV                              & 438                       & 0                        &  & GM                               & 18                       & 18                       \\
MD                                & 2,519                     & 0                        &  & LB                               & 34                       & 34                       \\
PM                                & 435                       & 0                        &  & LM                               & 349                      & 349                      \\ \cline{1-3}
\textbf{Total}                    & \textbf{4,920}            & \textbf{1,310}           &  & PM                               & 220                      & 220                      \\ \cline{1-3}
\multicolumn{3}{c}{\multirow{2}{*}{}}                                                    &  & YLC                              & 1,092                    & 1,092                    \\ \cline{5-7} 
\multicolumn{3}{c}{}                                                                     &  & \textbf{Total}                   & \textbf{3,353}           & \textbf{2,422}           \\ \cline{5-7} 
\end{tabular}
\end{threeparttable}
}
\end{table}

\subsection{Prerequisites for experimentation and evaluation}
In our experiment, each image belongs to a single class (i.e., no multiple infections). 
Therefore, for all the detection models, we follow the detection rule that if no bounding boxes are detected on an input image, the image is considered healthy, as mentioned above. 
The disease class is predicted based on the highest confidence among the predicted bounding boxes. 

For the two-stage model, given an input image, the CNN-based binary classifier is used in the first stage to determine whether it is healthy or diseased. 
Then, the baseline $\mathrm{M_{org}}$ model is applied in the second stage if the input is predicted as diseased. 
If no bounding boxes are detected after the second stage, that input is treated as healthy. 

\subsection{Training the detection-based models}
In this work, we fine-tuned all the detection models ($\mathrm{M_{org}},\mathrm{M_{HSM}},\mathrm{M_{HSReM}}$) from the YOLOv5-L6 model \citep{jocheryolov5}, which was pre-trained on the Microsoft COCO dataset \citep{lin2014microsoft}. 
The $\mathrm{M_{org}}$ model was trained only on the annotated disease classes (i.e., the Dataset A disease classes, excluding the healthy class). 
For the $\mathrm{M_{HSM}}$ and $\mathrm{M_{HSReM}}$ models, all disease images from the Dataset A training portion (Table \ref{tab:table_1}) and hard-samples from healthy cases (Table \ref{tab:table_2}) were used for training. 
In other words, $\mathrm{M_{HSM}}$ was trained with $16,705 + 4,920 = 21,625$ and $14,691 + 3,353 = 18,044$ images for cucumber and tomato, respectively. 
Similarly, $\mathrm{M_{HSReM}}$ was  trained with $16,705 + 1,310 = 18,015$ and $14,691 + 2,422 = 17,113$ images. 

All detection-based models were trained with an image size of $1,472\times 1,427$ pixels using the same hyperparameters and data augmentations as in the original YOLOv5 repository \citep{jocheryolov5}. 
During training, model checkpoints were regularly saved every five epochs, and the process was terminated after 100 epochs. 

\subsection{Training the classification-based models}
The CNN (EfficientNetV2) models were trained with class-level labels on Dataset A train (Table \ref{tab:table_1}) for both the tomato and cucumber datasets with the image size of $480\times480$ pixels. 
Note that this resolution was chosen because it was confirmed that using a higher resolution (e.g., $1,000\times 1,000$) would result in little performance improvement. 

The models were fine-tuned at all layers using the EfficientNetV2-Large \citep{tan2021efficientnetv2}, which was pre-trained on the ImageNet dataset \citep{deng2009imagenet} with the Adam optimizer \citep{kingma2015adam}. 
The mini-batch size was set to 32, and the learning rate was set at $10^{-4}$. 
Class weights sampling is used at training to handle the class imbalance. 
During training, we applied online augmentations, such as random rotations, flips, and blurring. 
Similar to the training of the detection-based models, checkpoints were saved at every five epochs, and the training is finished after 100 epochs. 

The same EfficientNetV2-Large model was used for the healthy/disease binary classifier in the first stage of the two-stage model and trained in the same way as the above CNN (EfficientNetV2) models. 
The difference is that all disease classes (Table \ref{tab:table_1}) were merged and treated as one disease class. 

\subsection{Classification results}
The first stage of the two-stage model (i.e., the CNN-based healthy binary classifiers) achieved 66.95\% and 90.70\% macro-average F1-scores for the cucumber and tomato test data (Dataset B), respectively. 
Those trained models were combined with the baseline ($\mathrm{M_{org}}$) models to form the two-stage models. 
\begin{figure*}[!t]
\centering
\begin{subfigure}[b]{0.32\textwidth}
    \centering
    \includegraphics[width=\textwidth]{figures/cucumber_naive_cropped.pdf}
    \caption{$\mathrm{M_{org}}$ (F1=74.71)}
\end{subfigure}
\begin{subfigure}[b]{0.32\textwidth}
    \centering
    \includegraphics[width=\textwidth]{figures/cucumber_HSM_cropped.pdf}
    \caption{$\mathrm{M_{HSM}}$ (F1=74.34)}
\end{subfigure}
\begin{subfigure}[b]{0.32\textwidth}
    \centering
    \includegraphics[width=\textwidth]{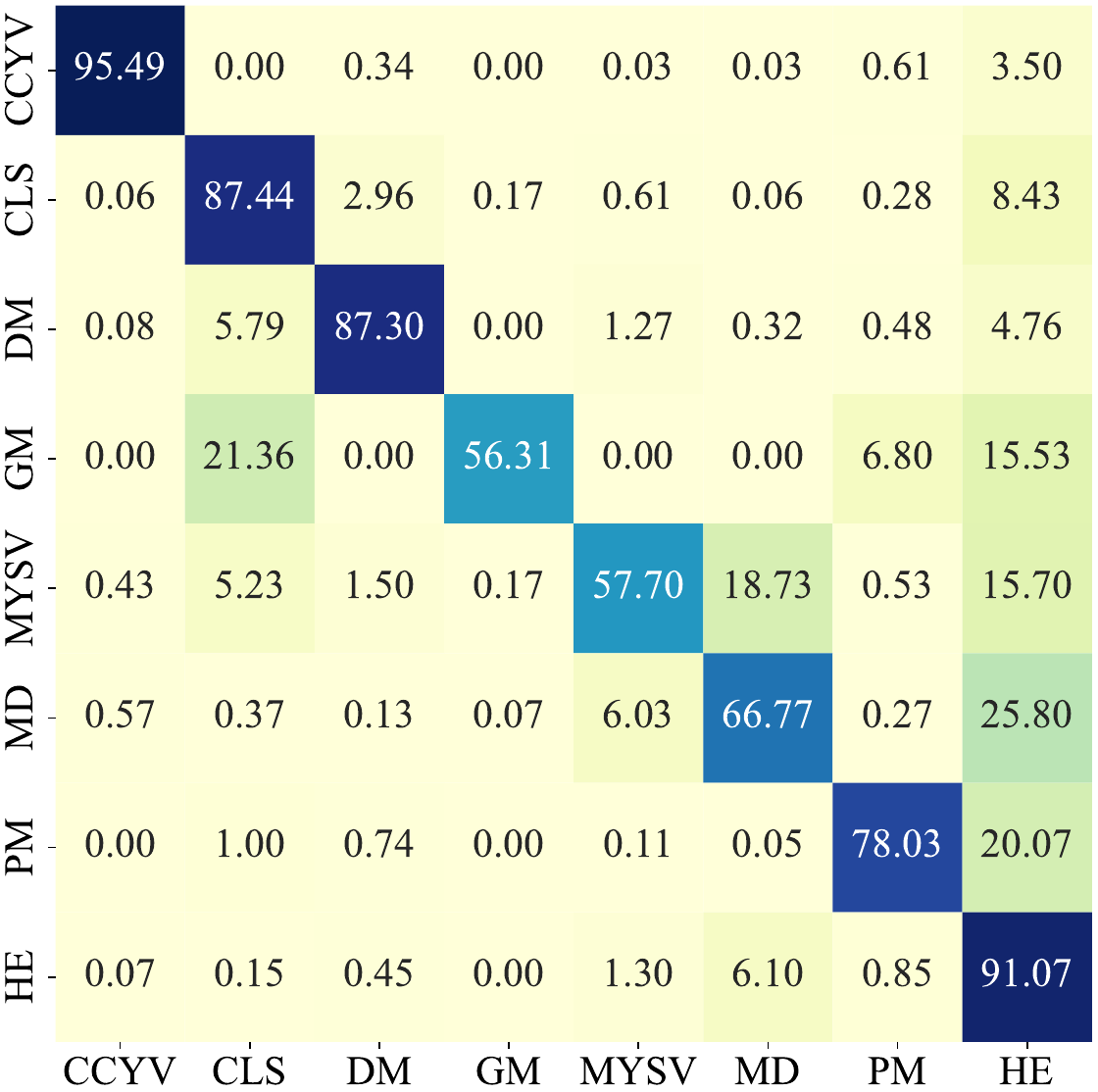}
    \caption{$\mathrm{M_{HSReM}}$ (F1=\textbf{79.76}) (proposed)}
\end{subfigure}
\begin{subfigure}[b]{0.32\textwidth}
    \centering
    \includegraphics[width=\textwidth]{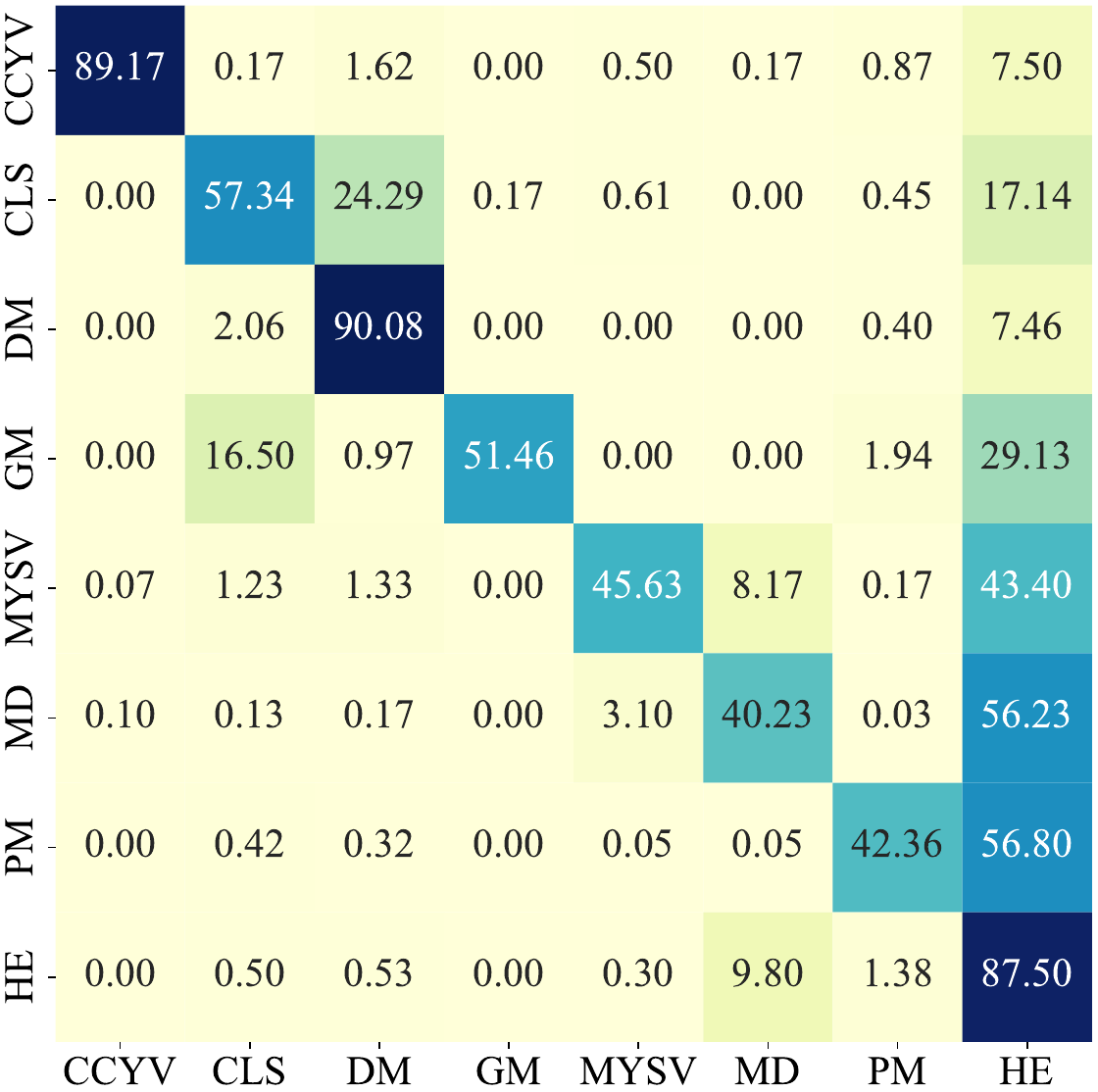}
    \caption{EfficientNetV2 (F1=66.62)}
\end{subfigure}
\begin{subfigure}[b]{0.32\textwidth}
    \centering
    \includegraphics[width=\textwidth]{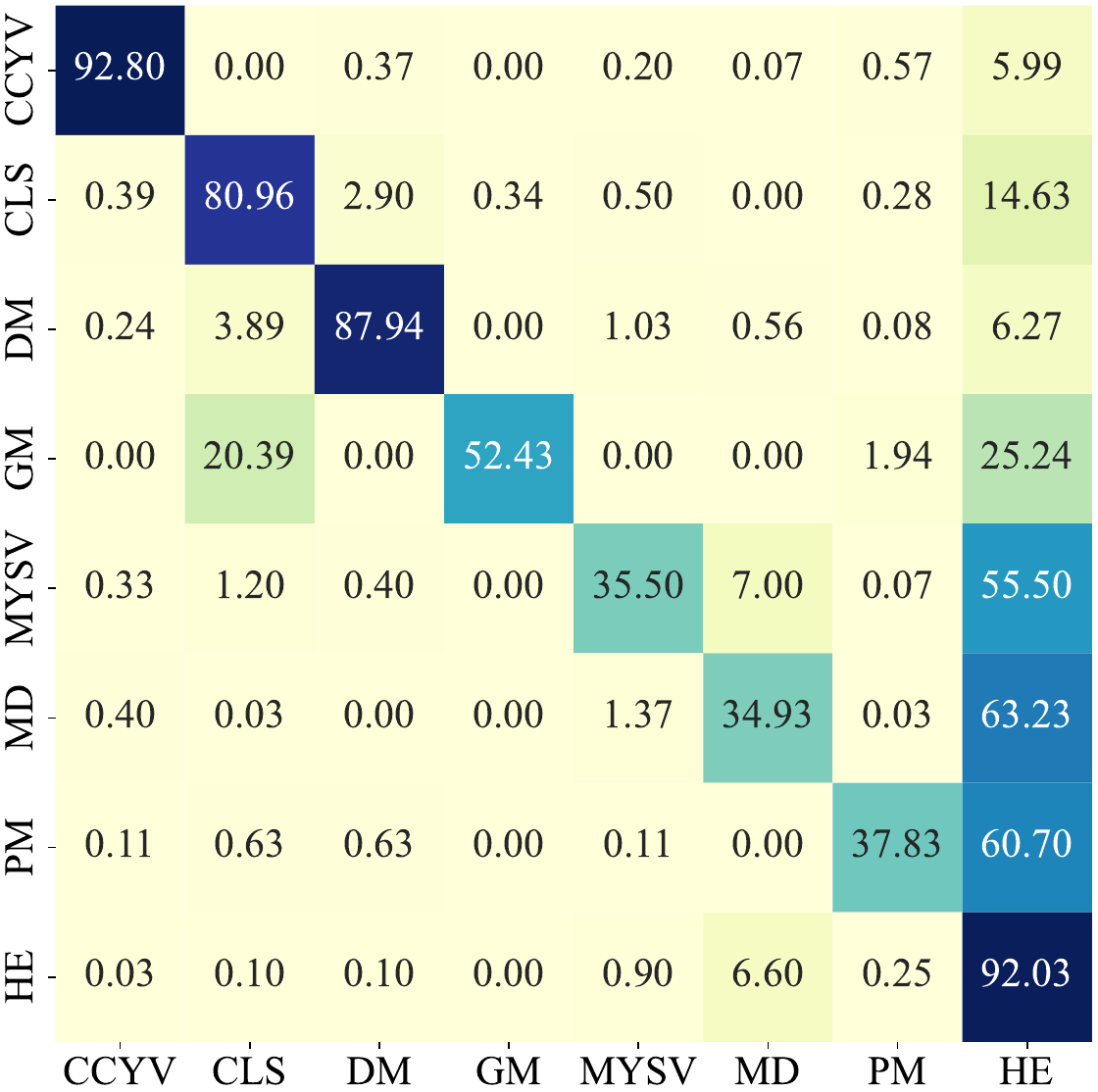}
    \caption{Two-stage (F1=68.32)}
\end{subfigure}
\caption{
    Confusion matrix comparisons (normalized over rows) of all models on the cucumber test data (Dataset B).
}
\label{fig:fig_5}
\end{figure*}
\begin{figure*}[!t]
\centering
\begin{subfigure}[b]{0.32\textwidth}
    \centering
    \includegraphics[width=\textwidth]{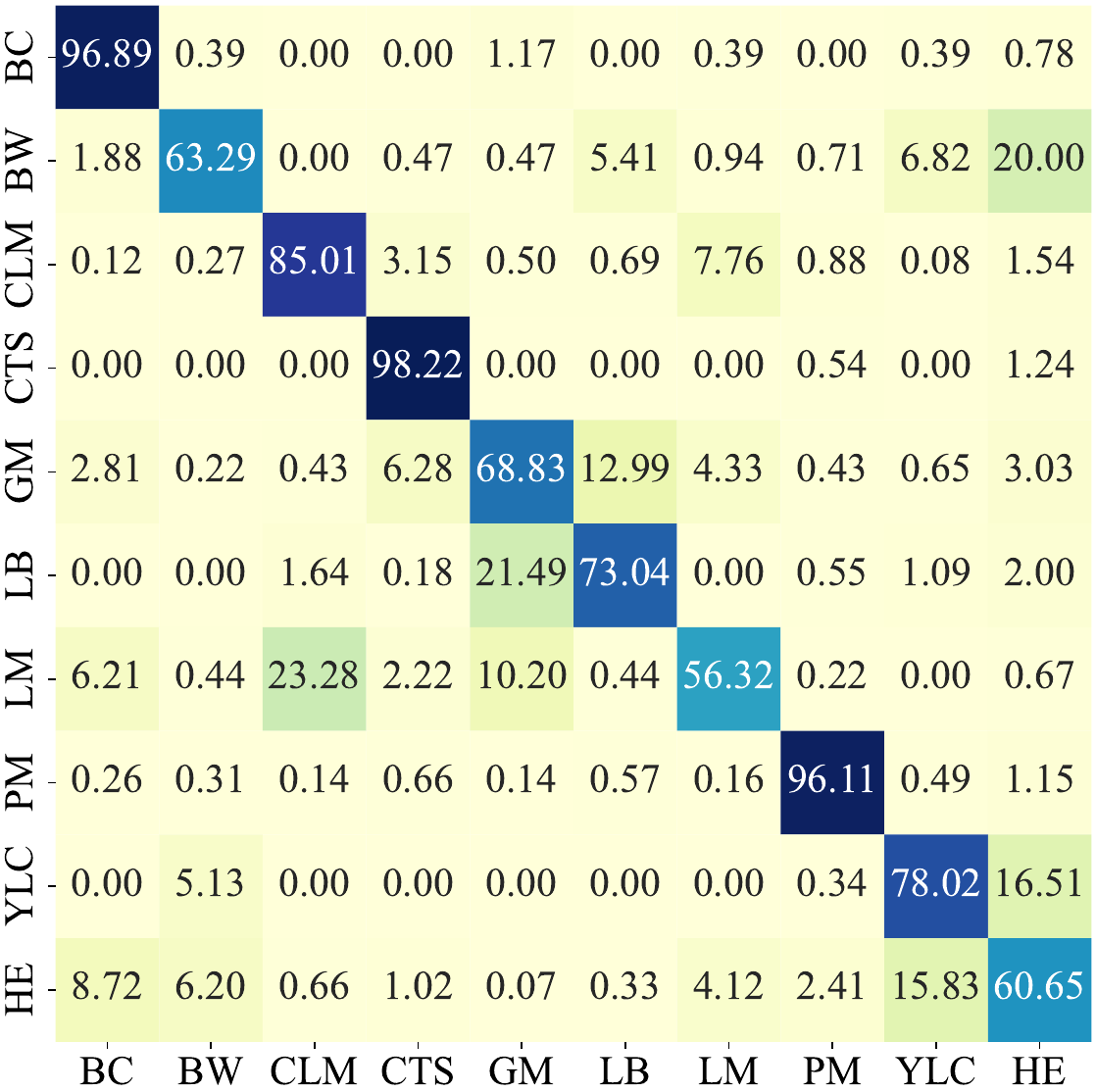}
    \caption{$\mathrm{M_{org}}$ (F1=72.19)}
\end{subfigure}
\begin{subfigure}[b]{0.32\textwidth}
    \centering
    \includegraphics[width=\textwidth]{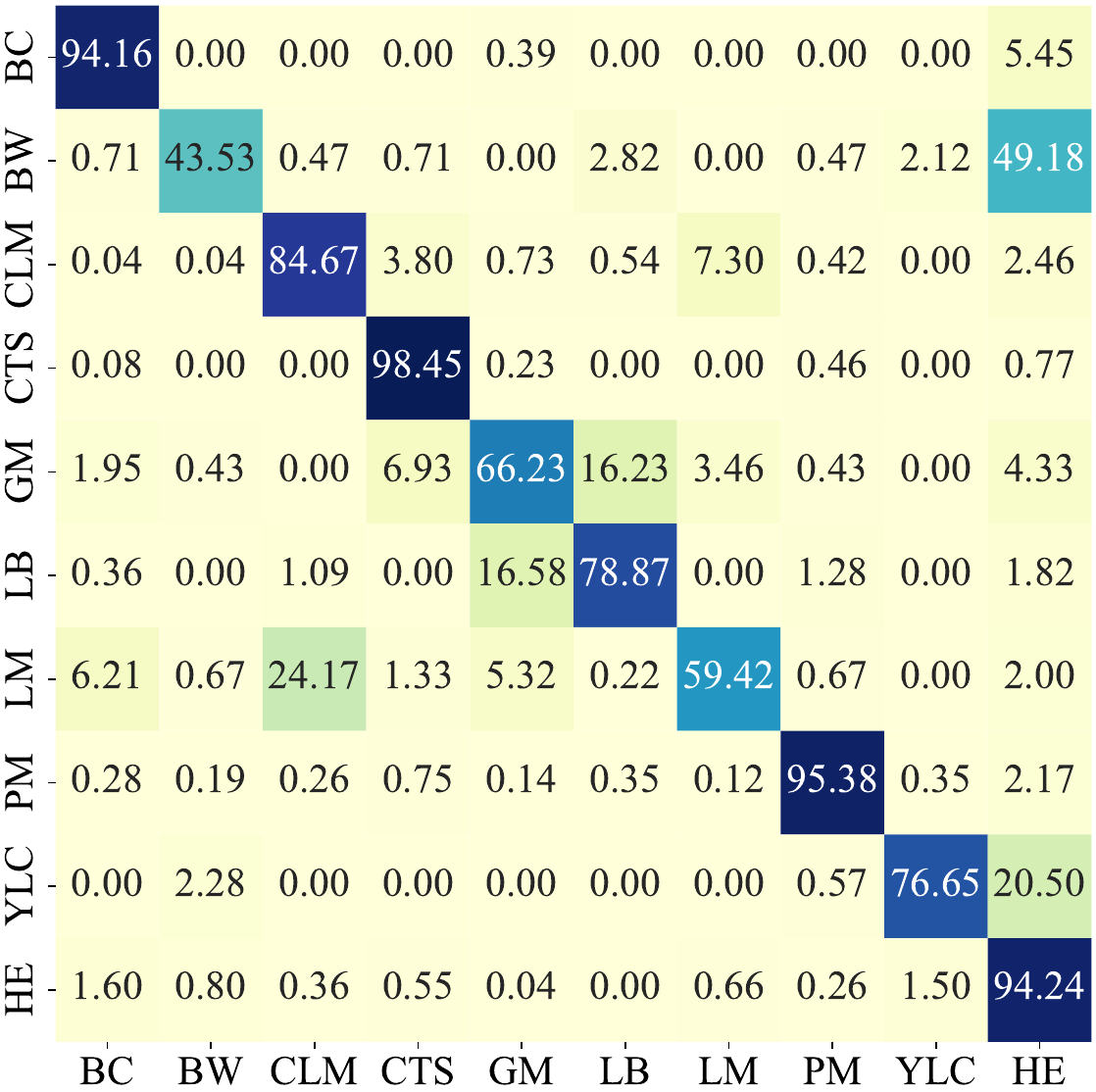}
    \caption{$\mathrm{M_{HSM}}$ (F1=78.79)}
\end{subfigure}
\begin{subfigure}[b]{0.32\textwidth}
    \centering
    \includegraphics[width=\textwidth]{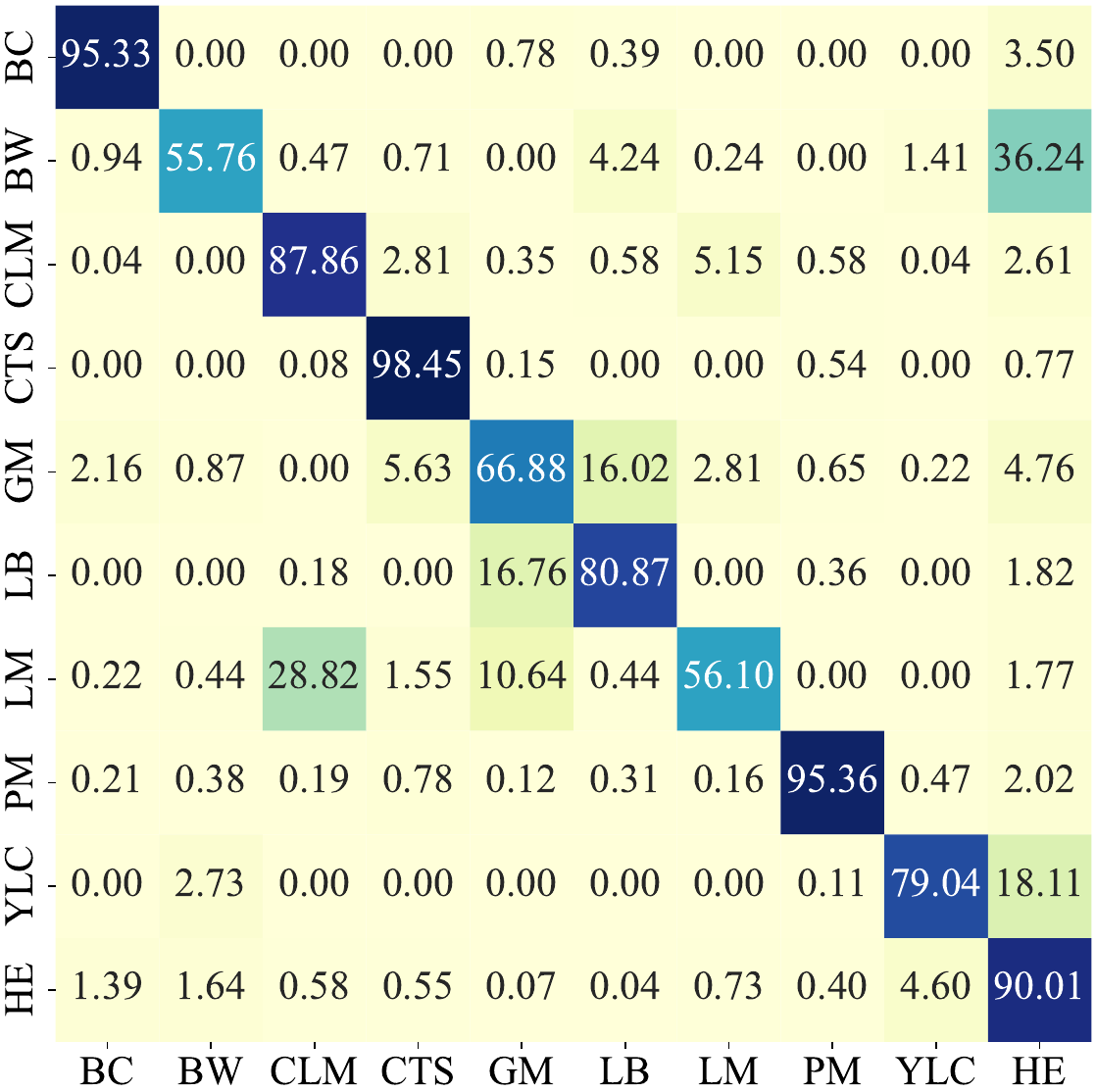}
    \caption{$\mathrm{M_{HSReM}}$ (F1=\textbf{80.09}) (proposed)}
\end{subfigure}
\begin{subfigure}[b]{0.32\textwidth}
    \centering
    \includegraphics[width=\textwidth]{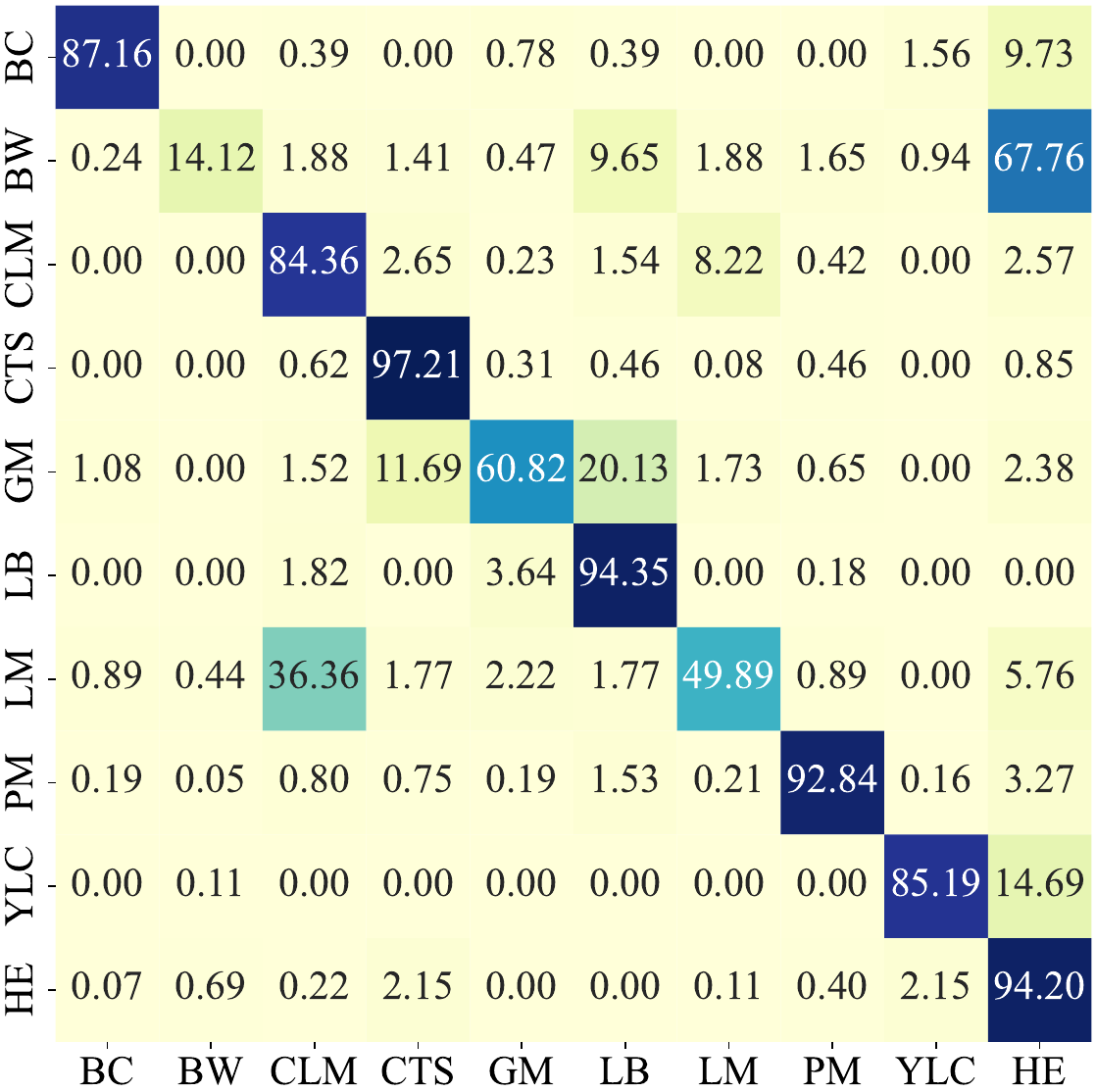}
    \caption{EfficientNetV2 (F1=75.83)}
\end{subfigure}
\begin{subfigure}[b]{0.32\textwidth}
    \centering
    \includegraphics[width=\textwidth]{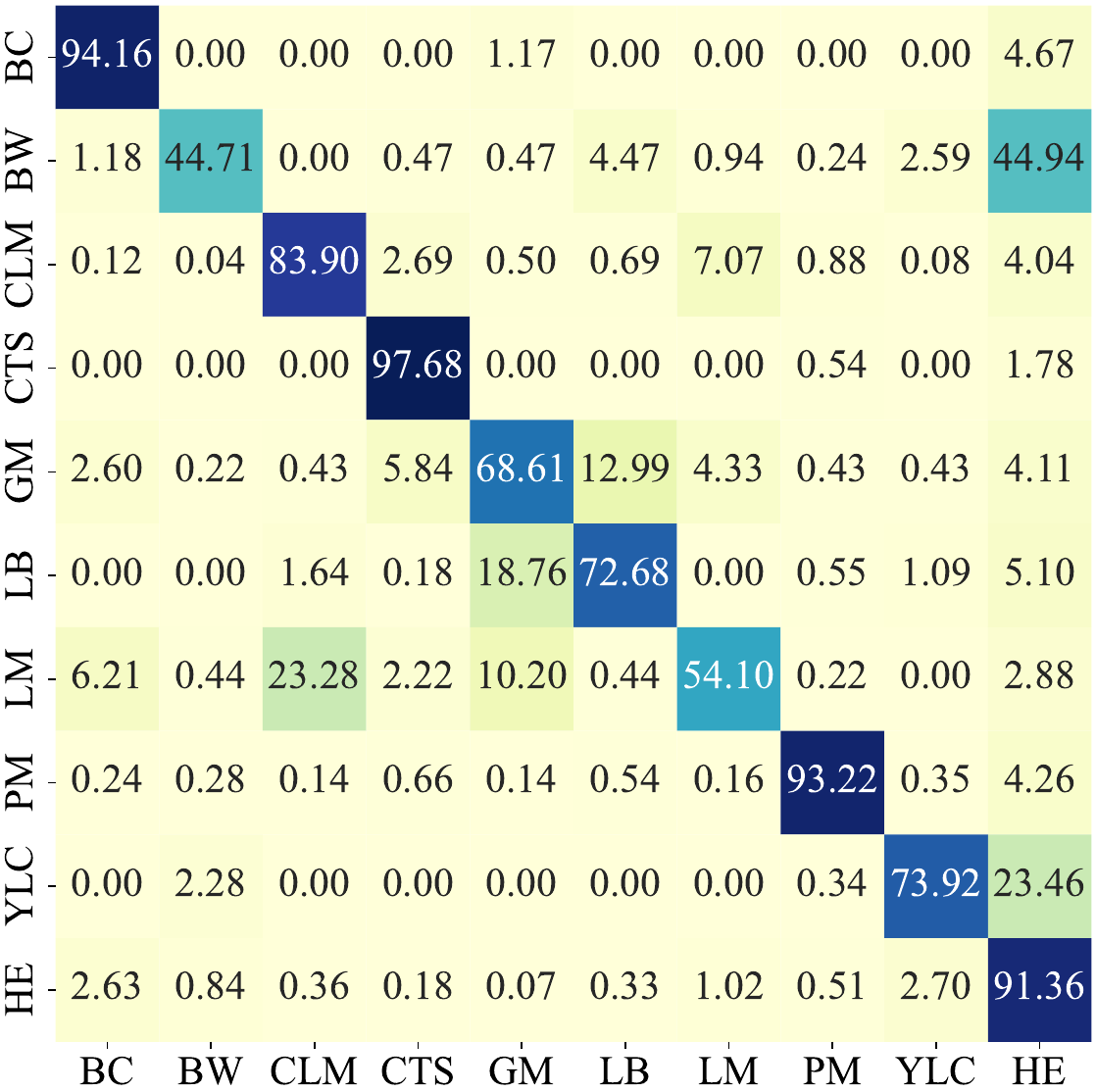}
    \caption{Two-stage (F1=76.64)}
\end{subfigure}
\caption{
    Confusion matrix comparisons (normalized over rows) of all models on tomato test data (Dataset B).
}
\label{fig:fig_6}
\end{figure*}

Figs. \ref{fig:fig_5} and \ref{fig:fig_6} show the comparison in terms of confusion matrices (normalized over rows), while Tables \ref{tab:table_3} and \ref{tab:table_4} summarize the diagnostic performance (in micro/macro F1-score and accuracy) on cucumber and tomato test data (Dataset B) among all models. 
There is a large gap between the micro-average F1-scores of Dataset A test and Dataset B. 
This problem is well-known as the covariate shift problem, and it typically occurs when the training and testing data are derived from different distributions \citep{ferentinos2018deep, cap2018, suwa2019comparable, saikawa2019aop, shibuya2022validation, wayama2023}. 
In general, our proposed baseline+HSReM ($\mathrm{M_{HSReM}}$) not only balances recall performance between the baseline ($\mathrm{M_{org}}$) and baseline+HSM ($\mathrm{M_{HSM}}$) but also achieved the best overall results. 
\begin{table*}[!t]
\centering
\captionof{table}{Classification results on cucumber data}
\label{tab:table_3}

\begin{subtable}{\textwidth}
\centering
\caption{Classification performance on Dataset A (train/test) and Dataset $\mathrm{B}^\ast$ in micro-average F1-score}

\resizebox{0.35\textwidth}{!}{
\begin{tabular}{lrrr}
\hline
                                                             & \multicolumn{2}{c}{Dataset A} &                                       \\ \cline{2-3}
\multirow{-2}{*}{Model}                                      & Train         & Test          & \multirow{-2}{*}{Dataset B}           \\ \hline
Baseline                                                     & 99.96         & 99.54         & 72.40                                 \\
+ HSM                                                        & 99.53         & 99.22         & \textbf{75.30}                                 \\
\begin{tabular}[c]{@{}l@{}}+ HSReM\\ (\textbf{proposed})\end{tabular} & 99.95         & 97.00         & {\color[HTML]{FF0000} \textbf{80.00}} \\
EfficientNetV2                                               & 99.96         & 99.30         & 65.16                                 \\
Two-stage                                                    & 99.96         & 99.57         & 65.92                                 \\ \hline
\end{tabular}
}
\end{subtable}

\vspace{1em} 

\begin{subtable}{\textwidth}
\centering
\caption{Per-class classification performance on Dataset $\mathrm{B}^\ast$ in F1-score and accuracy}

\resizebox{0.76\textwidth}{!}{
\begin{threeparttable}
\begin{tabular}{lrrrrrrrrr|r}
\hline
Model                                                        & CCYV                                  & CLS                                   & DM                                    & GM                                    & MYSV                                  & MD                                    & PM                                    & HE                                    & Avg.                                  & Acc.                                  \\ \hline
Baseline                                                     & 95.02                                 & {\color[HTML]{FF0000} \textbf{88.51}} & \textbf{88.26}                        & 66.29                                 & {\color[HTML]{FF0000} \textbf{69.50}} & 59.99                                 & {\color[HTML]{FF0000} \textbf{85.57}} & 44.58                                 & \textbf{74.71}                        & \textbf{62.21}                        \\
+ HSM                                                        & 95.34                                 & \textbf{87.54}                        & 87.86                                 & 53.99                                 & 58.80                                 & \textbf{64.01}                        & 77.28                                 & \textbf{69.91}                        & 74.34                                 & 59.25                                 \\
\begin{tabular}[c]{@{}l@{}}+ HSReM\\ (proposed)\end{tabular} & {\color[HTML]{FF0000} \textbf{97.11}} & 85.93                                 & 87.86                                 & {\color[HTML]{FF0000} \textbf{67.84}} & \textbf{69.32}                        & {\color[HTML]{FF0000} \textbf{68.88}} & \textbf{85.29}                        & {\color[HTML]{FF0000} \textbf{75.90}} & {\color[HTML]{FF0000} \textbf{79.76}} & {\color[HTML]{FF0000} \textbf{65.26}} \\
EfficientNetV2                                               & 94.19                                 & 69.98                                 & 76.92                                 & \textbf{66.67}                        & 60.83                                 & 49.77                                 & 57.35                                 & 57.28                                 & 66.62                                 & 50.37                                 \\
Two-stage                                                    & \textbf{95.68}                        & 86.21                                 & {\color[HTML]{FF0000} \textbf{90.12}} & 66.26                                 & 51.05                                 & 46.26                                 & 54.11                                 & 56.89                                 & 68.32                                 & 52.64                                 \\ \hline
\end{tabular}
\begin{tablenotes}[flushleft]\footnotesize
    \item[$\ast$] Images in Dataset B were captured in a distinct field from the training images (Dataset A).
    
    \item[$\dag$] Bold text in \textcolor{red}{\textbf{red}} and \textbf{black} indicate the best and second-best performance, respectively. \say{Acc.} indicates the accuracy.
\end{tablenotes}
\end{threeparttable}
}
\end{subtable}

\end{table*}
\begin{table*}[!t]
\centering
\captionof{table}{Classification results on tomato data}
\label{tab:table_4}

\begin{subtable}{\textwidth}
\centering
\caption{Classification performance on Dataset A (train/test) and Dataset $\mathrm{B}^\ast$ in micro-average F1-score}

\resizebox{0.35\textwidth}{!}{
\begin{tabular}{lrrr}
\hline
                                                             & \multicolumn{2}{c}{Dataset A} &                                       \\ \cline{2-3}
\multirow{-2}{*}{Model}                                      & Train         & Test          & \multirow{-2}{*}{Dataset B}           \\ \hline
Baseline                                                     & 99.56         & 99.10         & 82.00                                 \\
+ HSM                                                        & 99.53         & 99.05         & \textbf{87.86}                        \\
\begin{tabular}[c]{@{}l@{}}+ HSReM\\ (proposed)\end{tabular} & 99.69         & 97.09         & {\color[HTML]{FF0000} \textbf{88.15}} \\
EfficientNetV2                                               & 99.44         & 98.04         & 86.54                                 \\
Two-stage                                                    & 99.41         & 98.90         & 85.94                                 \\ \hline
\end{tabular}
}
\end{subtable}

\vspace{1em} 

\begin{subtable}{\textwidth}
\centering
\caption{Per-class classification performance on Dataset $\mathrm{B}^\ast$ in F1-score and accuracy}

\resizebox{0.86\textwidth}{!}{
\begin{threeparttable}
\begin{tabular}{lrrrrrrrrrrr|r}
\hline
Model                                                        & BC                                    & BW                                    & CLM                                   & CTS                                   & GM                                    & LB                                    & LM                                    & PM                                    & YLC                                   & HE                                    & Avg.                                  & Acc.                                  \\ \hline
Baseline                                                     & 61.63                                 & \textbf{57.66}                        & \textbf{89.30}                        & 92.59                                 & 65.57                                 & 73.85                                 & 48.29                                 & 96.76                                 & 66.54                                 & 69.73                                 & 72.19                                 & 64.43                                 \\
+ HSM                                                        & 80.80                                 & 55.56                                 & 89.14                                 & 92.47                                 & \textbf{67.03}                        & \textbf{78.80}                        & \textbf{56.54}                        & \textbf{97.13}                        & \textbf{83.29}                        & {\color[HTML]{FF0000} \textbf{87.09}} & \textbf{78.79}                        & \textbf{66.71}                        \\
\begin{tabular}[c]{@{}l@{}}+ HSReM\\ (proposed)\end{tabular} & \textbf{86.73}                        & {\color[HTML]{FF0000} \textbf{62.95}} & {\color[HTML]{FF0000} \textbf{90.61}} & \textbf{93.49}                        & 66.38                                 & {\color[HTML]{FF0000} \textbf{79.50}} & {\color[HTML]{FF0000} \textbf{57.57}} & {\color[HTML]{FF0000} \textbf{97.17}} & 80.42                                 & \textbf{86.05}                        & {\color[HTML]{FF0000} \textbf{80.09}} & {\color[HTML]{FF0000} \textbf{68.77}} \\
EfficientNetV2                                               & {\color[HTML]{FF0000} \textbf{89.42}} & 23.58                                 & 87.19                                 & 90.48                                 & {\color[HTML]{FF0000} \textbf{70.69}} & 78.43                                 & 48.97                                 & 95.79                                 & {\color[HTML]{FF0000} \textbf{88.00}} & 85.80                                 & 75.83                                 & 62.45                                 \\
Two-stage                                                    & 76.95                                 & 56.38                                 & 88.79                                 & {\color[HTML]{FF0000} \textbf{93.58}} & 66.46                                 & 73.96                                 & 52.03                                 & 95.86                                 & 79.29                                 & 83.15                                 & 76.64                                 & 64.63                                 \\ \hline
\end{tabular}
\begin{tablenotes}[flushleft]\footnotesize
    \item[$\ast$] Images in Dataset B were captured in a distinct field from the training images (Dataset A).
    
    \item[$\dag$] Bold text in \textcolor{red}{\textbf{red}} and \textbf{black} indicate the best and second-best performance, respectively. \say{Acc.} indicates the accuracy.
\end{tablenotes}
\end{threeparttable}
}
\end{subtable}

\end{table*}

\section{Discussion}
    \subsection{Disease diagnosis comparisons}
Based on Tables \ref{tab:table_3} and \ref{tab:table_4}, because the baseline ($\mathrm{M_{org}}$) models were not trained on any healthy data, their performance for healthy plants is lowest, as expected. 
In contrast, training with all healthy hard-samples, as in baseline+HSM ($\mathrm{M_{HSM}}$), boosted the performance for the healthy class but degraded the diagnostic performance for other diseases with similar appearances to healthy plants (see Fig. \ref{fig:fig_2}). 
Thanks to the proposed HSReM strategy, with the HSS module for strategically selecting an appropriate level of hard-samples, our baseline+HSReM ($\mathrm{M_{HSReM}}$) models successfully delivered robust diagnostic performance for the healthy and diseased classes, achieving the best overall results on both large-scale cucumber and tomato unseen test data among detection-based models. 

In comparison with the object detection-based models, the CNN-based model (EfficientNetV2) scored the lowest for cucumbers, below the baseline ($\mathrm{M_{org}}$) model. 
This is because, unlike the object detection models, the CNN could not distinguish between cucumbers with fine-grained, small lesions (CLS, DM, MD, and PM) and healthy ones, even though it had been trained on many healthy data. 
Notably, the overall diagnostic performance of our proposed $\mathrm{M_{HSReM}}$ model is superior to the EfficientNetV2 model, being 13.14 and 14.89 points better in macro-average F1-score and accuracy, respectively (Table \ref{tab:table_3}). 
For tomatoes, the CNN-based model performed better than the $\mathrm{M_{org}}$ model. 
This was because the proportion of small lesions was lower in tomatoes than in cucumbers. However, there were still some lesions with extremely low discrimination rates (BW and LM), making the model less than practical. 
It is noteworthy that detection-based models are beneficial from from high-resolution data (we used $1,472\times 1,472$ as input images) but the CNN-based models are unlikely to derive similar benefits. 
In our preliminary experiments, training CNN-based disease classification models on $1,000\times 1,000$ pixel data does not help to improve the performance as compared to the model with $480\times480$ pixel input (as in the EfficientNetV2 models). 

The two-stage models seemed to be a promising approach when dealing with healthy data. 
However, their overall performance heavily depended on the first stage (the CNN-based binary classifier). 
Given the abovementioned covariate shift problem, as well as the high similarity between health and disease, the overall disease diagnostic performance was significantly affected. 

\subsection{The effect of adding healthy data on diagnostic performance}
\begin{figure*}[!t]
\centering
\includegraphics[width=0.95\textwidth]{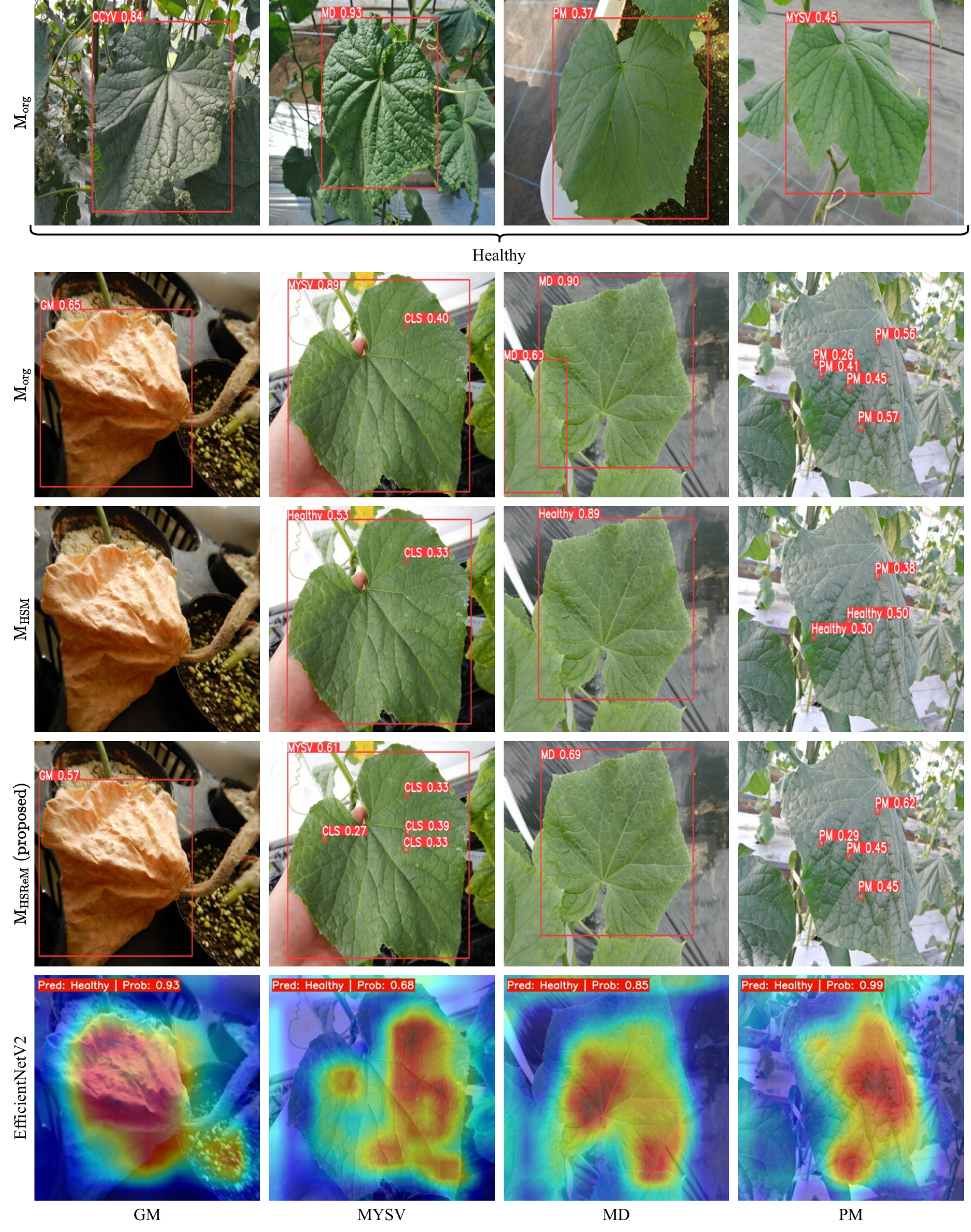}
\caption{
    The visual detection comparison on cucumber test data (Dataset B). 
    The Grad-CAM++ \citep{chattopadhay2018grad} is used to visualize explanations of the EfficientNetV2 model predictions in the form of heatmaps (last row). 
    The warmer color of the heatmap, the more it contributes to the predicted class.
}
\label{fig:fig_7}
\end{figure*}

We observed that the negative effect of adding healthy data to training was most evident in the cucumber case (Fig. \ref{fig:fig_5} and Table \ref{tab:table_3}) because there are many viral disease classes that are very visually similar to healthy data (see Fig. \ref{fig:fig_2}). 
Fig. \ref{fig:fig_7} shows a visual comparison of the detection results on cucumber test data. 
The top row shows examples of $\mathrm{M_{org}}$'s false positive detection of healthy cases, and the rows beneath are detection examples for the $\mathrm{M_{org}}$, $\mathrm{M_{HSM}}$, $\mathrm{M_{HSReM}}$, and EfficientNetV2 models on disease cases. 
The results of the EfficientNetV2 model are represented as heatmaps by using GradCAM++ \citep{chattopadhay2018grad}. 

Because the $\mathrm{M_{org}}$ model was not trained on any healthy images, it falsely detected healthy images as diseases (top row), while its diagnosis performance on diseases was still robust (second row). 
Without the careful selection of healthy hard-samples for training, the $\mathrm{M_{HSM}}$ model failed to precisely diagnose disease images, due to the strong similarity between healthy and diseased data (third row). 
This means that simply adding more data does not always improve overall disease diagnostic performance. 
This also made the performance of the two-stage model, which is limited by the performance of the first-stage discriminator, less than desirable. 

With the appropriate number of training hard-samples having been selected by the HSReM strategy, the $\mathrm{M_{HSReM}}$ model showed robustness in many difficult disease cases (fourth row). 
Note that the EfficientNetV2 model could not detect fine-grained symptoms and, thus, failed to diagnose disease in those cases. 

In the tomato case, the $\mathrm{M_{HSM}}$ model had significantly improved performance as compared to the baseline $\mathrm{M_{org}}$ model (Fig. \ref{fig:fig_6} and Table \ref{tab:table_4}), and the negative impact of adding healthy data in this case is much smaller than in the case of cucumber. 
This is because tomato leaves with disease symptoms are quite distinguishable from healthy ones. 
Most such diseases are not viral diseases and have fairly obvious symptoms. 
The high performance of the CNN-based healthy binary classifier used in the two-stage model also supports this. 
Reviewing the results in detail, the addition of healthy data, even in the HSM model for tomatoes, improved the diagnostic performance as compared to the baseline $\mathrm{M_{org}}$ model for many diseases, including the BC class. 
However, we argue that this is only the case when disease symptoms are distinguishable from healthy ones. 

Here, the proposed HSReM selected all but BW among the hard-samples obtained via HSM. 
Therefore, the overall results are similar to but better than those of HSM. 
In particular, for BW, which performed poorly at the baseline (F1 = 57.66) and even worse in HSM (F1 = 55.56), HSReM achieved significant performance improvement (F1 = 62.95), confirming the effectiveness of the proposed method. 

For tomatoes, for which it was relatively easy to discriminate between healthy cases and disease classes, the two-stage model showed some success. 
However, for diseases that were difficult to diagnose with the baseline model, performance was limited as compared to that achieved via HSM and the proposed HSReM. 
These results support the idea that the proposed method can perform well in tasks such as diagnosing plant diseases that are extremely difficult to identify.

\subsection{Limitations}
Despite the HSReM strategy having demonstrated favorable results, some limitations remain. First, applying the HSReM strategy requires training both the $\mathrm{M_{org}}$ and $\mathrm{M_{HSM}}$ models. 
Thus, this process is somewhat time-consuming. 
Although hardware performance has improved and it has become much easier to train object detection models, further transforming this process into an end-to-end process will also be valuable. 
Second, the current HSReM strategy still requires a certain number of hard-sample training data to achieve the desired results. 
In extreme cases in which there are few hard-samples remaining after applying the HSS module, the diagnostic performance will be equivalent to that of the $\mathrm{M_{org}}$ model. 
In this case, if collecting more hard-samples is challenging, generating more leaf samples by applying certain generative-based methods \citep{cap2020leafgan,kanno2021ppig} is a promising approach. 
Third, currently, the selection of the threshold $\theta$ in the HSS module is performed manually, and the reduction rate is fixed to apply in all classes. 
We believe that the efficiency of the HSReM strategy can be further improved by developing algorithms that better control the threshold $\theta$ and reduction rate on a class-by-class basis. 
There is potential to enhance our proposal, and we intend to investigate these areas in future work.

\section{Conclusion}
    In this paper, we have proposed a simple but very practical training strategy called hard-sample re-mining (HSReM) for detection-based plant disease diagnosis systems to use in dealing with newly added healthy data. 
With the capability to strategically select the optimal quantity of training data, our HSReM strategy not only effectively preserves the model’s performance on healthy data but also improves the disease diagnosis performance on other unseen disease data. 
We firmly believe that the HSReM strategy is a practical and robust approach that can have a substantial impact on automated crop disease diagnosis. 
    
\section*{Acknowledgment}
This work was supported by the Ministry of Agriculture, Forestry and Fisheries of Japan (MAFF) commissioned project study on the development of pest diagnosis technology using AI (JP17935051) and by the Cabinet Office, Public/Private R\&D Investment Strategic Expansion Program (PRISM). We would like to express our sincere thanks to all the experts and test site personnel who actually grew the plants in their respective fields, strictly controlled pests and diseases, and took an extremely large number of high-quality photographs in conducting this study.

\nocite{*}
\footnotesize{
\bibliographystyle{IEEEtran}
\bibliography{reference}
}

\end{document}